\documentclass{article} 
\usepackage{iclr2021_conference}
\usepackage{times}
\usepackage{graphicx} 
\usepackage{hyperref}
\usepackage{url}
\usepackage{enumitem}
\usepackage{subcaption}
\usepackage{booktabs}
\usepackage[toc,page]{appendix}
\usepackage[T1]{fontenc}
\usepackage[scaled=.8]{beramono}
\usepackage{sidecap}
\usepackage{multirow}
\usepackage{placeins}
\usepackage{algpseudocode}
\usepackage{algorithm}
\usepackage{amssymb}

\usepackage{amsmath,amsfonts,bm}

\def\eqref#1{equation~\ref{#1}}

\def\1{\bm{1}}

\DeclareMathAlphabet{\mathsfit}{\encodingdefault}{\sfdefault}{m}{sl}
\SetMathAlphabet{\mathsfit}{bold}{\encodingdefault}{\sfdefault}{bx}{n}

\newcommand*\samethanks[1][\value{footnote}]{\footnotemark[#1]}

\title{Asymmetric self-play for automatic goal discovery in robotic manipulation}

\author{%
    OpenAI \\
    Matthias Plappert\thanks{Authors are listed at random and a detailed contribution section is at the end. Please cite as \textbf{OpenAI et al.}. }, 
    Raul Sampedro\samethanks,  
    Tao Xu\samethanks, 
    Ilge Akkaya\samethanks, 
    Vineet Kosaraju\samethanks, 
    Peter Welinder\samethanks,\\
    Ruben D'Sa\samethanks, 
    Arthur Petron\samethanks, 
    Henrique Ponde de Oliveira Pinto\samethanks, 
    Alex Paino\samethanks, 
    Hyeonwoo Noh\samethanks, \\
    Lilian Weng\samethanks, 
    Qiming Yuan\samethanks, 
    Casey Chu\samethanks, 
    Wojciech Zaremba\samethanks\\
}

\iclrfinalcopy 
\begin{document}

\abovedisplayskip=-2pt
\abovedisplayshortskip=-2pt
\belowdisplayskip=5pt
\belowdisplayshortskip=5pt
\abovecaptionskip=5pt
\belowcaptionskip=-2pt

\maketitle

\begin{abstract}

We train a single, goal-conditioned policy that can solve many robotic manipulation tasks, including tasks with previously unseen goals and objects. We rely on asymmetric self-play for goal discovery, where two agents, Alice and Bob, play a game. Alice is asked to propose challenging goals and Bob aims to solve them. We show that this method can discover highly diverse and complex goals without any human priors.
Bob can be trained with only sparse rewards, because the interaction between Alice and Bob results in a natural curriculum and Bob can learn from Alice's trajectory when relabeled as a goal-conditioned demonstration.  
Finally, our method scales, resulting in a single policy that can generalize to many unseen tasks such as setting a table, stacking blocks, and solving simple puzzles.
Videos of a learned policy is available at \url{https://robotics-self-play.github.io}.

\end{abstract}

\section{Introduction}
\label{sec:intro}
We are motivated to train a \emph{single} goal-conditioned policy~\citep{kaelbling1993learning} that can solve \emph{any} robotic manipulation task that a human may request in a given environment.
In this work, we make progress towards this goal by solving a robotic manipulation problem in a table-top setting where the robot's task is to change the initial configuration of a variable number of objects on a table to match a given goal configuration.
This problem is simple in its formulation but likely to challenge a wide variety of cognitive abilities of a robot as objects become diverse and goals become complex.

\begin{figure*}[b!]
    \vspace{-0.3cm}
    \centering
    \begin{subfigure}[b]{0.38\textwidth}
        \centering
        \includegraphics[height=3.2cm]{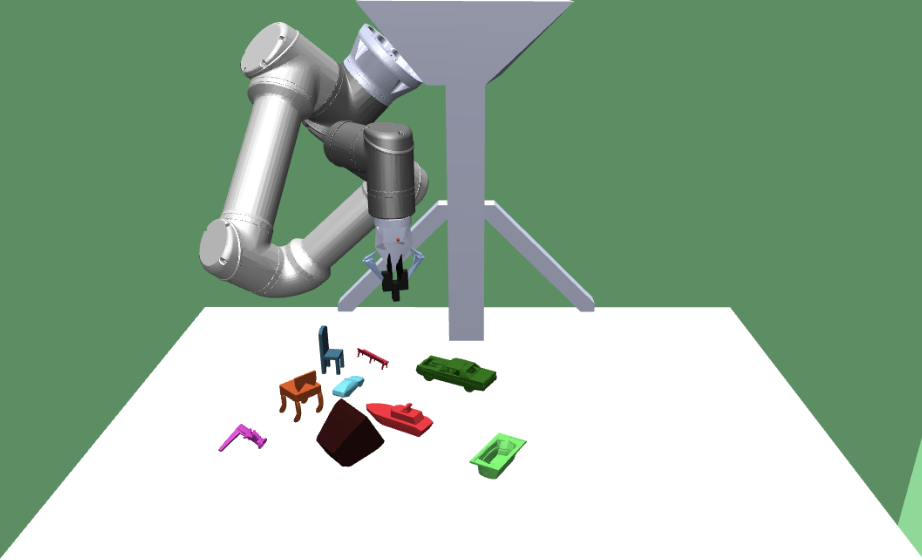}
        \caption{\label{fig:table_top}Table-top setting with a robot arm}
    \end{subfigure}
    \begin{subfigure}[b]{0.32\textwidth}
        \centering
        \includegraphics[height=3.2cm]{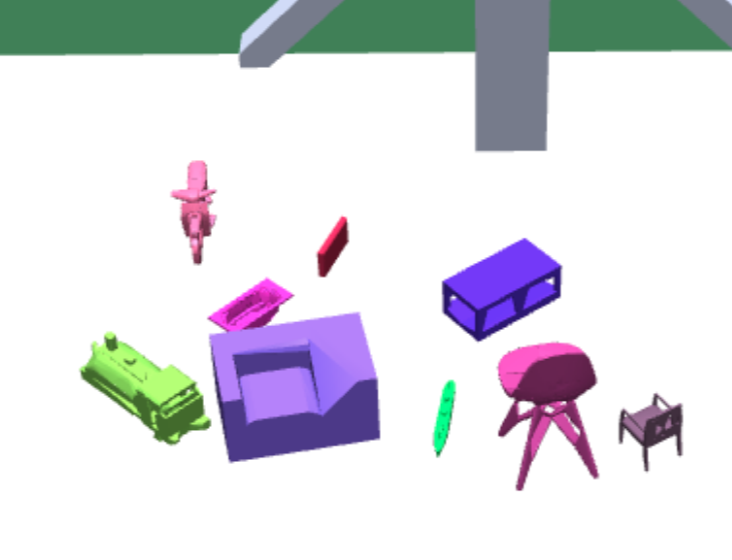}
        \caption{\label{fig:init_state}Example initial state for training}
    \end{subfigure}
    \begin{subfigure}[b]{0.28\textwidth}
        \centering
        \includegraphics[height=3.2cm]{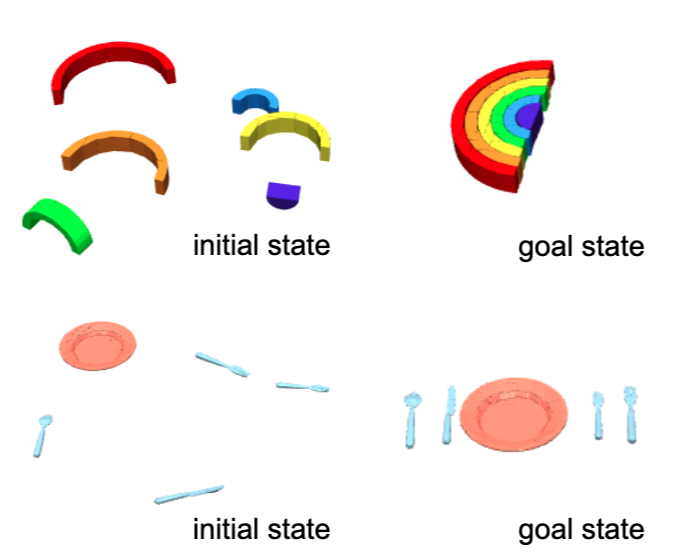}
        \caption{\label{fig:holdout_two_states}Example holdout tasks}
    \end{subfigure}
    \caption{\label{fig:train_env}(a) We train a policy that controls a robot arm operating in a table-top setting. (b) Randomly placed ShapeNet~\citep{chang2015shapenet} objects constitute an initial state distribution for training. (c) We use multiple manually designed holdout tasks to evaluate the learned policy.}
\end{figure*}

Motivated by the recent success of deep reinforcement learning for robotics~\citep{levine2016end,gu2017deep,hwangbo2019learning,openai2019solving}, we tackle this problem using deep reinforcement learning on a very large training distribution.
An open question in this approach is how we can build a training distribution rich enough to achieve generalization to many unseen manipulation tasks.
This involves defining both an environment’s initial state distribution and a goal distribution.
The initial state distribution determines how we sample a set of objects and their configuration at the beginning of an episode, and the goal distribution defines how we sample target states given an initial state.
In this work, we focus on a scalable way to define a rich goal distribution.




The research community has started to explore automated ways of defining goal distributions. For example, previous works have explored learning a generative model of goal distributions~\citep{florensa2017goalGAN,nair2018RIG,racaniere2019setter} and collecting teleoperated robot trajectories to identify goals~\citep{lynch2020play,gupta2019relay}.
In this paper, we extend an alternative approach called asymmetric self-play~\citep{sukhbaatar2017intrinsic, sukhbaatar2018learning} for automated goal generation.
Asymmetric self-play trains two RL agents named Alice and Bob.
Alice learns to propose goals that Bob is likely to fail at, and Bob, a goal-conditioned policy, learns to solve the proposed goals.
Alice proposes a goal by manipulating objects and Bob has to solve the goal starting from the same initial state as Alice's.
By embodying these two agents into the same robotic hardware, this setup ensures that all proposed goals are provided with at least one solution: Alice's trajectory.

There are two main reasons why we consider asymmetric self-play to be a promising goal generation and learning method. First, any proposed goal is \emph{achievable}, meaning that there exists at least one solution trajectory that Bob can follow to achieve the goal. Because of this property, we can exploit Alice's trajectory to provide additional learning signal to Bob via behavioral cloning. This additional learning signal alleviates the overhead of heuristically designing a curriculum or reward shaping for learning. 
Second, this approach does not require labor intensive data collection.

In this paper, we show that asymmetric self-play can be used to train a goal-conditioned policy for complex object manipulation tasks, and the learned policy can zero-shot generalize to many manually designed holdout tasks, which consist of either previously unseen goals, previously unseen objects, or both.
To the best of our knowledge, this is the first work that presents zero-shot generalization to many previously unseen tasks by training purely with asymmetric self-play.\footnote{Asymmetric self-play is proposed in~\citet{sukhbaatar2017intrinsic, sukhbaatar2018learning}, but to supplement training while the majority of training is conducted on target tasks. Zero-shot generalization to unseen tasks was not evaluated.}

\section{Problem Formulation}
\label{sec:formulation}
\begin{figure*}
    \centering
    \begin{subfigure}[b]{0.34\textwidth}
        \centering
        \includegraphics[width=\textwidth]{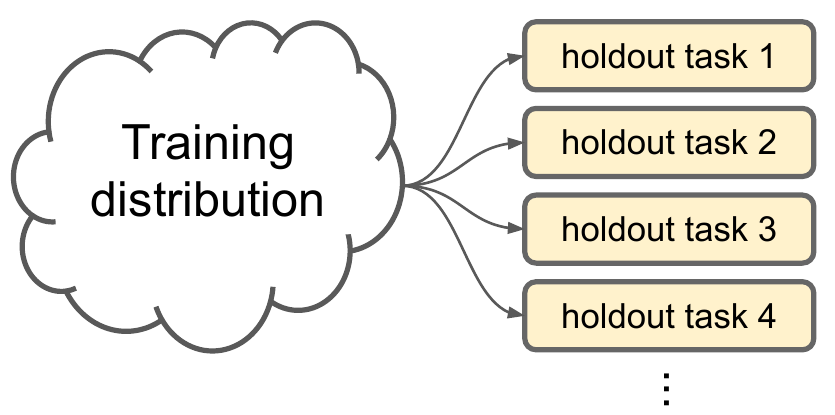}
        \caption{\label{fig:framework}Overall framework}
    \end{subfigure}
    \begin{subfigure}[b]{0.65\textwidth}
        \centering
        \includegraphics[width=\textwidth]{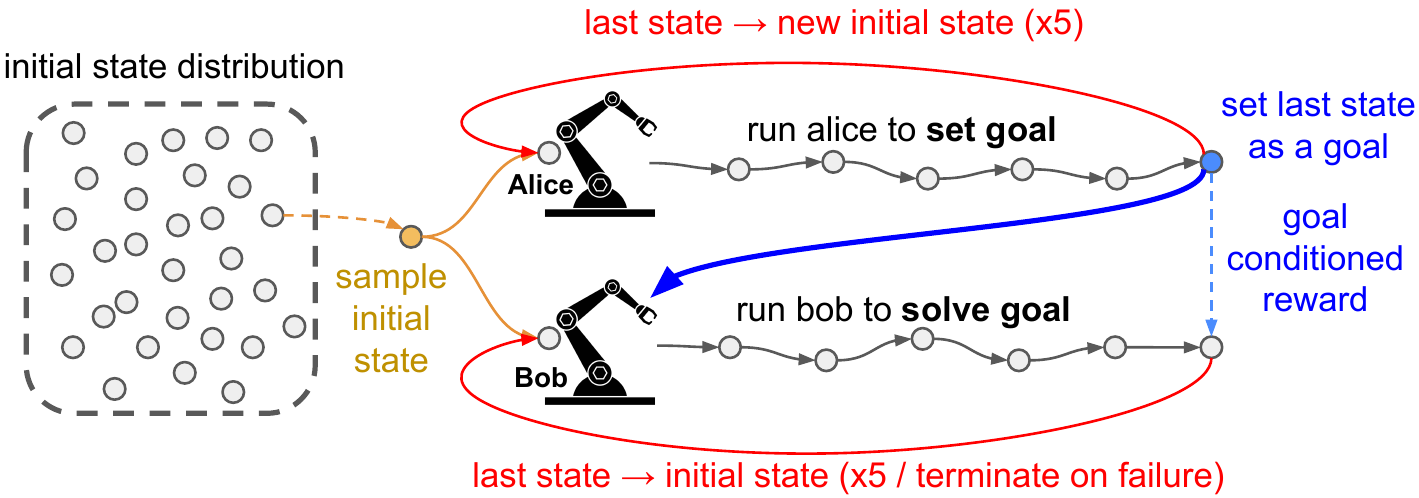}
        \caption{\label{fig:framework_selfplay}Training distribution based on asymmetric self-play}
    \end{subfigure}
    \caption{(a) We train a goal-conditioned policy on a single training distribution and evaluate its performance on many unseen holdout tasks. (b) To construct a training distribution, we sample an initial state from a predefined distribution, and run a goal setting policy (Alice) to generate a goal. In one episode, Alice is asked to generate 5 goals and Bob solves them in sequence until it fails.\vspace{-0.8em}}
\end{figure*}

Our training environment for robotic manipulation consists of a robot arm with a gripper attached and a wide range of objects placed on a table surface (Figure~\ref{fig:table_top},\ref{fig:init_state}). The goal-conditioned policy learns to control the robot to rearrange randomly placed objects (the initial state) into a specified goal configuration (\autoref{fig:holdout_two_states}).
%
We aim to train a policy on a single training distribution and to evaluate its performance over a suite of holdout tasks which are independently designed and not explicitly present during training (\autoref{fig:framework}).
In this work, we construct the training distribution via \emph{asymmetric self-play} (\autoref{fig:framework_selfplay}) to achieve generalization to many unseen holdout tasks~(\autoref{fig:holdout_two_states}).

\paragraph{Mathematical formulation} Formally, we model the interaction between an environment and a goal-conditioned policy as a goal-augmented Markov decision process $\mathcal{M} = \langle \mathcal{S}, \mathcal{A}, \mathcal{P}, \mathcal{R}, \mathcal{G} \rangle$, where $\mathcal{S}$ is the state space, $\mathcal{A}$ is the action space, $\mathcal{P}:\mathcal{S}\times\mathcal{A}\times\mathcal{S}\mapsto\mathbb{R}$ denotes the transition probability, $\mathcal{G} \subseteq \mathcal{S}$ specifies the goal space and $\mathcal{R}:\mathcal{S}\times\mathcal{G}\mapsto\mathbb{R}$ is a goal-specific reward function. 
A goal-augmented trajectory sequence is $\{(s_0, g, a_0, r_0), \dots, (s_t, g, a_t, r_t) \}$, where the goal is provided to the policy as part of the observation at every step.
We say a goal is achieved if $s_t$ is sufficiently close to $g$ (Appendix~\ref{appendix:reward}). 
With a slightly overloaded notation, we define the \emph{goal distribution} $\mathcal{G}(g \vert s_0)$ as the probability of a goal state $g \in \mathcal{G}$ conditioned on an initial state $s_0 \in \mathcal{S}$.

\paragraph{Training goal distribution} 
A naive design of the goal distribution $\mathcal{G}(g \vert s_0)$ is to randomly place objects uniformly on the table, but it is unlikely to generate interesting goals, such as an object picked up and held above the table surface by a robot gripper. Another possible approach, collecting tasks and goals manually, is expensive and hard to scale.
We instead sidestep these issues and automatically generate goals via training based on asymmetric self-play~\citep{sukhbaatar2017intrinsic, sukhbaatar2018learning}. Asymmetric self-play involves using a policy named Alice $\pi_A(a \vert s)$ to set goals and a goal-conditioned policy Bob $\pi_B(a \vert s, g)$ to solve goals proposed by Alice, as illustrated in \autoref{fig:framework_selfplay}. 
We run $\pi_A$ to generate a trajectory $\tau_A = \{(s_0, a_0, r_0), \dots, (s_T, a_T, r_T)\}$ and the last state is labelled as a goal $g$ for $\pi_B$ to solve. The goal distribution $\mathcal{G}(s_T=g \vert s_0)$ is fully determined by $\pi_A$
%
and we train Bob only on this goal distribution. We therefore say \emph{zero-shot generalization} when Bob generalizes to a holdout task which is not explicitly encoded into the training distribution.

\paragraph{Evaluation on holdout tasks} To assess zero-shot generalization of $\pi_B(a\vert s, g)$ from our training setup, we hand-designed a suite of holdout tasks with goals that are never directly incorporated into the training distribution. Some holdout tasks also feature previously unseen objects.
The holdout tasks are designed to either test whether a specific skill has been learned, such as the ability to pick up objects (\autoref{fig:block_skills}), or represent a semantically interesting task, such as setting a table (\autoref{fig:holdout_two_states}).
Appendix~\ref{appendix:holdouts} describes the list of holdout tasks that we use in our experiments.
Note that none of the holdout tasks are used for training $\pi_B(a\vert s, g)$.

\section{Asymmetric Self-play}
\label{sec:methods}

To train Alice policy $\pi_A(a|s)$ and Bob policy $\pi_B(a|s,g)$, we run the following multi-goal game within one episode, as illustrated in \autoref{fig:framework_selfplay}:
\begin{enumerate}
    \item An initial state $s_0$ is sampled from an initial state distribution. Alice and Bob are instantiated into their own copies of the environment. Alice and Bob alternate turns as follows.
    \item \textbf{Alice's turn.} Alice interacts with its environment for a fixed number of $T$ steps and may rearrange the objects. The state at the end of Alice's turn $s_T$ will be used as a goal $g$ for Bob. If the proposed goal is invalid (e.g.~if Alice has not moved any objects, or if an object has fallen off the table), the episode terminates.
    \item \textbf{Bob's turn.} Bob receives reward if it successfully achieves the goal $g$ in its environment. Bob's turn ends when it succeeds at achieving the goal or reaches a timeout. If Bob's turn ends in a failure, its remaining turns are skipped and treated as failures, while we let Alice to keep generating goals.
    \item Alice receives reward if Bob fails to solve the goal that Alice proposed. Steps 2--3 are repeated until 5 goals are set by Alice or Alice proposes an invalid goal, and then the episode terminates.
\end{enumerate}

The competition created by this game encourages Alice to propose goals that are increasingly challenging to Bob, while Bob is forced to solve increasingly complex goals. The multi-goal setup was chosen to allow Bob to take advantage of environmental information discovered earlier in the episode to solve its remaining goals, which \citet{openai2019solving} found to be important for transfer to physical systems. Note however that in this work we focus on solving goals in simulation only.
To improve stability and avoid forgetting, we have Alice and Bob play against past versions of their respective opponent in $20\%$ of games.
More details about the game structure and pseudocode for training with asymmetric self-play are available in Appendix~\ref{appendix:game}.

\subsection{Reward Structure}

For Bob, we assign \emph{sparse} goal-conditioned rewards. 
We measure the positional and rotational distance between an object and its goal state as the Euclidean distance and the Euler angle rotational distance, respectively. Whenever both distance metrics are below a small error (the \emph{success threshold}), this object is deemed to be placed close enough to the goal state and Bob receives 1 reward immediately. But if this object is moved away from the goal state that it has arrived at in past steps, Bob obtains -1 reward such that the sum of per-object reward is at most 1 during a given turn.
When all of the objects are in their goal state, Bob receives 5 additional reward and its turn is over.

For Alice, we assign a reward after Bob has attempted to solve the goal: 5 reward if Bob failed at solving the goal, and 0 if Bob succeeded.
We shape Alice's reward slightly by adding 1 reward if it has set a valid goal, defined to be when no object has fallen off the table and any object has been moved more than the success threshold. An additional penalty of $-3$ reward is introduced when Alice sets a goal with objects outside of the placement area, defined to be a fixed 3D volume within the view of the robot's camera. More details are discussed in Appendix~\ref{appendix:reward}.

\subsection{Alice Behavioral Cloning (ABC)}
\label{sec:abc}

One of the main benefits of using asymmetric self-play is that the generated goals come with at least one solution to achieve it: \emph{Alice's trajectory}.
Similarly to~\cite{sukhbaatar2018learning}, we exploit this property by training Bob with Behavioral Cloning (BC) from Alice's trajectory, in addition to the reinforcement learning (RL) objective. We call this learning mechanism \emph{Alice Behavioral Cloning} (ABC). We propose several improvements over the original formulation in~\cite{sukhbaatar2018learning}.



\paragraph{Demonstration trajectory filtering} Compared to BC from expert demonstrations, using Alice's trajectory needs extra care.
Alice's trajectory is likely to be suboptimal for solving the goal, as Alice might arrive at the final state merely by accident. Therefore, we only consider trajectories with goals that Bob failed to solve as demonstrations, to avoid distracting Bob with suboptimal examples. Whenever Bob fails, we relabel Alice's trajectory $\tau_A$ to be a goal-augmented version $\tau_\text{BC} = \{(s_0, s_T, a_0, r_0), \dots, (s_T, s_T, a_T, r_T)\}$ as a demonstration for BC, where $s_T$ is the goal.

\paragraph{PPO-style BC loss clipping} The objective for training Bob is $\mathcal{L} = \mathcal{L}_\text{RL} + \beta \mathcal{L}_\text{abc}$, where $\mathcal{L}_\text{RL}$ is an RL objective and $\mathcal{L}_\text{abc}$ is the ABC loss. $\beta$ is a hyperparameter controlling the relative importance of the BC loss. We set $\beta=0.5$ throughout the whole experiment.
A naive BC loss is to minimize the negative log-likelihood of demonstrated actions,  $- \mathbb{E}_{(s_t, g_t, a_t) \in \mathcal{D}_\text{BC}} \big[ \log\pi_B(a_t \vert s_t, g_t; \theta) \big]$ where $\mathcal{D}_\text{BC}$ is a mini-batch of demonstration data and $\pi_B$ is parameterized by $\theta$. We found that overly-aggressive policy changes triggered by BC sometimes led to learning instabilities.
To prevent the policy from changing too drastically, we introduce PPO-style loss clipping~\citep{schulman2017ppo} on the BC loss by setting the advantage $\hat{A}=1$ in the clipped surrogate objective:

\begin{equation*}
    \mathcal{L}_\text{abc} = - \mathbb{E}_{(s_t, g_t, a_t) \in \mathcal{D}_\text{BC}} \bigg[\text{clip}\Big(\frac{\pi_B(a_t \vert s_t, g_t; \theta)}{\pi_B(a_t \vert s_t, g_t; \theta_\text{old})}, 1 - \epsilon, 1 + \epsilon \Big)\bigg]
    \label{eq:abc_loss}
\end{equation*}

where $\pi_B(a_t \vert s_t, g_t; \theta)$ is Bob's likelihood on a demonstration based on the parameters that we are optimizing, and  $\pi_B(a_t\vert s_t, g_t; \theta_\text{old})$ is the likelihood based on Bob's behavior policy (at the time of demonstration collection) evaluated on a demonstration. This behavior policy is identical to the policy that we use to collect RL trajectories. 
By setting $\hat{A}=1$, this objective optimizes the naive BC loss, but clips the loss whenever $\frac{\pi_B(a_t \vert s_t, g_t; \theta)}{\pi_B(a_t\vert s_t, g_t; \theta_\text{old})}$ is bigger than $1+\epsilon$, to prevent the policy from changing too much.
$\epsilon$ is a clipping threshold and we use $\epsilon=0.2$ in all the experiments. 

\section{Related Work}
\label{sec:background}

\textbf{Training distribution for RL}
In the context of multi-task RL~\citep{beattie2016deepmind,hausman2018learning,yu2020meta}, multi-goal RL~\citep{kaelbling1993learning,andrychowicz2017hindsight}, and meta RL~\citep{wang2016metaRL,duan2016rl2}, previous works manually designed a distribution of tasks or goals to see better generalization of a policy to a new task or goal.
Domain randomization~\citep{sadeghi2016cad2rl,tobin2017domain,openai2020block} manually defines a distribution of simulated environments, but in service of generalizing to the same task in the real world.


There are approaches to grow the training distribution automatically~\citep{srivastava2013powerplay}.
Self-play~\citep{tesauro1995tdgammon,silver2016alphago,silver2017zero,bansal2017selfplay,openai2019dota,vinyals2019starcraft} constructs an ever-growing training distribution where multiple agents learn by competing with each other, so that the resulting agent shows strong performance on a single game.
\cite{openai2019solving} automatically grew a distribution of domain randomization parameters to accomplish better generalization in the task of solving a Rubik's cube on the physical robot.
\cite{wang2019poet,wang2020poet2} studied an automated way to keep discovering challenging 2D terrains and locomotion policies that can solve them in a 2D bipedal walking environment.

We employ asymmetric self-play to construct a training distribution for learning a goal-conditioned policy and to achieve generalization to unseen tasks.
\citet{florensa2017goalGAN,nair2018RIG,racaniere2019setter} had the same motivation as ours, but trained a generative model instead of a goal setting policy. Thus, the difficulties of training a generative model were inherited by these methods: difficulty of modeling a high dimensional space and generation of unrealistic samples.
\citet{lynch2020play,gupta2019relay} used teleoperation to collect arbitrary robot trajectories, and defined a goal distribution from the states in the collected trajectories. This approach likely requires a large number of robot trajectories for each environment configuration (e.g.~ various types of objects on a table), and randomization of objects was not studied in this context.

\textbf{Asymmetric self-play} Asymmetric self-play was proposed by~\cite{sukhbaatar2017intrinsic} as a way to supplement RL training. 
\cite{sukhbaatar2017intrinsic} mixed asymmetric self-play training with standard RL training on the target task and measured the performance on the target task. 
\cite{sukhbaatar2018learning} used asymmetric self-play to pre-train a hierarchical policy and evaluated the policy after fine-tuning it on a target task.
\cite{liu2019cer} adopted self-play to encourage efficient learning with sparse reward in the context of an exploration competition between a pair of agents.
As far as we know, no previous work has trained a goal-conditioned policy \textit{purely} based on asymmetric self-play and evaluated generalization to unseen holdout tasks.

\textbf{Curriculum learning}
Many previous works showed the difficulty of RL and proposed an automated curriculum~\citep{andrychowicz2017hindsight,florensa2017reverse,salimans2018backward,matiisen2019TSCL,zhang2020automatic} or auxiliary exploration objectives~\citep{oudeyer2007lp,baranes2013motivation,pathak2017curiosity,burda2018exploration,ecoffet2019GoExplore,ecoffet2020first} to learn \emph{predefined tasks}.
When training goal-conditioned policies, relabeling or reversing trajectories~\citep{andrychowicz2017hindsight,florensa2017reverse,salimans2018backward} or imitating successful demonstrations~\citep{oh2018SIL,ecoffet2019GoExplore,ecoffet2020first} naturally reduces the task complexity.
Our work shares a similarity in that asymmetric self-play alleviates the difficulty of learning a goal-conditioned policy via an intrinsic curriculum and imitation from the goal setter's trajectory, but our work does not assume any predefined task or goal distribution.

\textbf{Hierarchical reinforcement learning (HRL)}
Some HRL methods jointly trained a goal setting policy (high-level or manager policy) and a goal solving policy (low-level or worker policy)~\citep{vezhnevets2017feudal,levy2019HAC,nachum2018HIRO}.
However, the motivation for learning a goal setting policy in HRL is not to challenge the goal solving policy, but to cooperate to tackle a task that can be decomposed into a sequence of sub-goals.
Hence, this goal setting policy is trained to optimize task reward for the target task, unlike asymmetric self-play where the goal setter is rewarded upon the other agent's failure.

\textbf{Robot learning for object manipulation.} It has been reported that training a policy for multi-object manipulation is very challenging with \emph{sparse} rewards~\citep{riedmiller2018sacX,vecerik2018robotics}.
One example is block stacking, which has been studied for a long time in robotics as it involves complex contact reasoning and long horizon motion planning~\citep{deisenroth2011stacking}. Learning block stacking often requires a hand-designed curriculum~\citep{li2019multiobj}, meticulous reward shaping~\citep{popov2017lego}, fine-tuning~\citep{rusu2017progressive}, or human demonstrations~\citep{nair2018demo,duan2017oneshot}.
%
In this work, we use block stacking as one of the holdout tasks to test zero-shot generalization, but without training on it.


\section{Experiments}
\label{sec:experiments}

In this section, we first show that asymmetric self-play generates an effective training curriculum that enables generalization to unseen hold-out tasks.
Then, the experiment is scaled up to train in an environment containing multiple random complex objects and evaluate it with a set of holdout tasks containing unseen objects and unseen goal configurations. Finally, we demonstrate how critical ABC is for Bob to make progress in a set of ablation studies.

\subsection{Experimental setup}

\begin{SCfigure}[50][t!]
\caption{\label{fig:block_skills}Holdout tasks in the environment using 1 or 2 blocks. The transparent blocks denote the desired goal state, while opaque blocks are the current state. (a) \texttt{push}: The blocks must be moved to their goal locations and orientations. There is no differentiation between the six block faces. (b) \texttt{flip}: Each side of the block is labelled with a unique letter. The blocks must be moved to make every face correctly positioned as what the goal specifies. (c) \texttt{pick-and-place}: One goal block is in the air. (d) \texttt{stack}: Two blocks must be stacked in the right order at the right location.
}
\includegraphics[width=5.5cm]{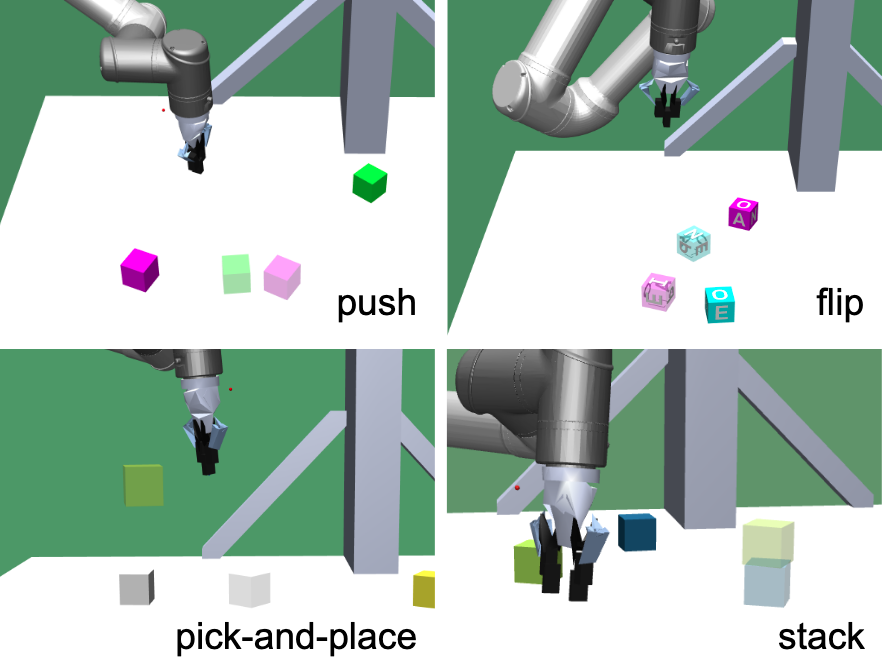}
\end{SCfigure}

We implement the training environment\footnote{Our training and evaluation environments are publicly available at \url{https://github.com/openai/robogym}} described in Sec.~\ref{sec:formulation} with randomly placed ShapeNet objects~\citep{chang2015shapenet} as an initial state distribution. In addition, we set up another simpler environment using one or two blocks of fixed size, used for small-scale comparisons and ablation studies.
\autoref{fig:block_skills} visualizes four holdout tasks for this environment.
Each task is designed to evaluate whether the robot has acquired certain manipulation skills: pushing, flipping, picking up and stacking blocks. 
Experiments in Sec.~\ref{sec:exp-curriculum}, \ref{sec:exp-train-dist} and \ref{sec:exp-ablation} focus on blocks and experimental results based on ShapeNet objects are present on Sec.~\ref{sec:big-exp}.
More details on our training setups are in Appendix~\ref{appendix:setup}.

We implement Alice and Bob as two independent policies of the same network architecture with memory (Appendix~\ref{appendix:model_arch}), except that Alice has no observation on goal state.
The policies take state observations (``state policy'') for experiments with blocks (Sec. \ref{sec:exp-curriculum}, \ref{sec:exp-train-dist}, and \ref{sec:exp-ablation}), and take both vision and state observations (``hybrid policy'') for experiments with ShapeNet objects (Sec.~\ref{sec:big-exp}).
Both policies are trained with Proximal Policy Optimization (PPO)~\citep{schulman2017ppo}.

\subsection{Generalization to unseen goals without manual curricula}
\label{sec:exp-curriculum}

One way to train a single policy to acquire all the skills in \autoref{fig:block_skills} is to train a goal-conditioned policy directly over a mixture of these tasks. However, training directly over these tasks without a curriculum turns out to be very challenging, as the policy completely fails to make any progress.\footnote{The tasks was easier when we ignored object rotation as part of the goal, and used a smaller table.}
In contrast, Bob is able to solve all these holdout tasks quickly when learning via asymmetric self-play, without explicitly encoding any prior knowledge of the holdout tasks into the training distribution.

\begin{figure}[t!]
\centering
\includegraphics[width=\linewidth]{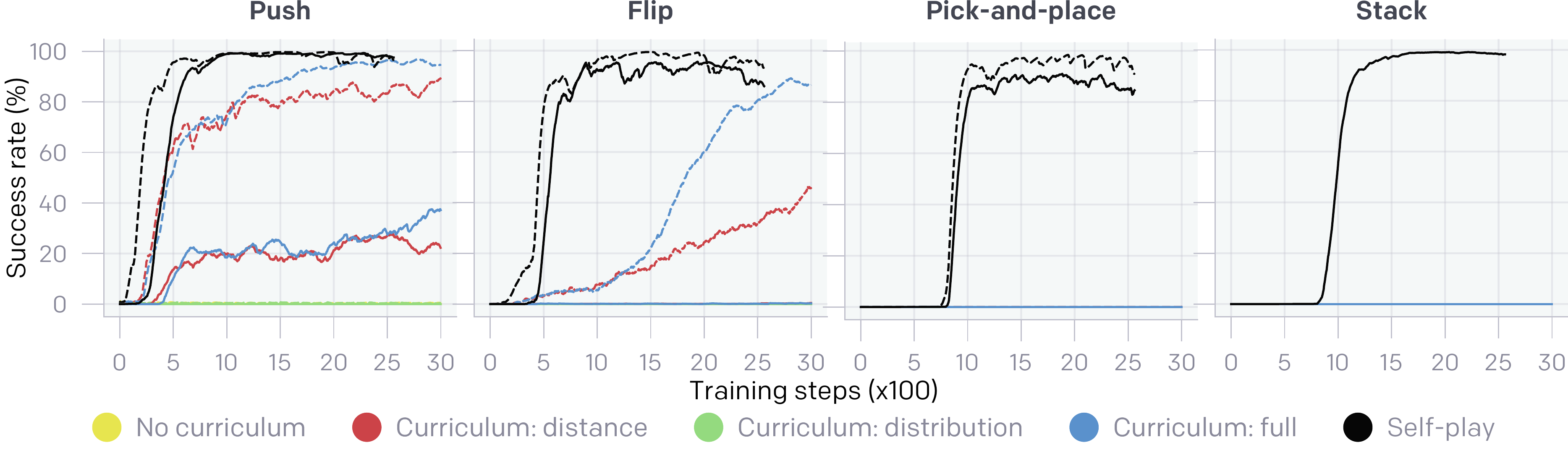}
\caption{\label{fig:exp-curriculum}
Generalization to unseen holdout tasks for blocks.
Baselines are trained over a mixture of all holdout tasks.
The solid lines represent 2-blocks, while the dashed lines are for 1-block.
The x-axis denotes the number of training steps via asymmetric self-play.
The y-axis is the zero-shot generalization performance of Bob policy at corresponding training checkpoints.
Note that success rate curves of completely failed baselines are occluded by others.
}
\end{figure}

To gauge the effect of an intrinsic curriculum introduced by self-play, we carefully designed a set of non-self-play baselines using explicit curricula controlled by Automatic Domain Randomization~\citep{openai2019solving}.
All baselines are trained over a mixture of block holdout tasks as the goal distribution. 
We measure the effectiveness of a training setup by tracking the success rate for each holdout task, as shown in \autoref{fig:exp-curriculum}. 
The \texttt{no curriculum} baseline fails drastically. 
The \texttt{curriculum:distance} baseline expands the distance between the initial and goal states gradually as training progresses, but only learns to push and flip a single block. 
The \texttt{curriculum:distribution} baseline, which slowly increases the proportion of pick-and-place and stacking goals in the training distribution, fails to acquire any skill. 
The \texttt{curriculum:full} baseline incorporates all hand-designed curricula yet still cannot learn how to pick up or stack blocks. We have spent a decent amount of time iterating and improving these baselines but found it especially difficult to develop a scheme good enough to compete with asymmetric self-play. See Appendix~\ref{appendix:baselines-curriculum} for more details of our baselines.


\subsection{Discovery of novel goals and solutions}
\label{sec:exp-train-dist}

\begin{figure}[t!]
\centering
\includegraphics[width=\linewidth]{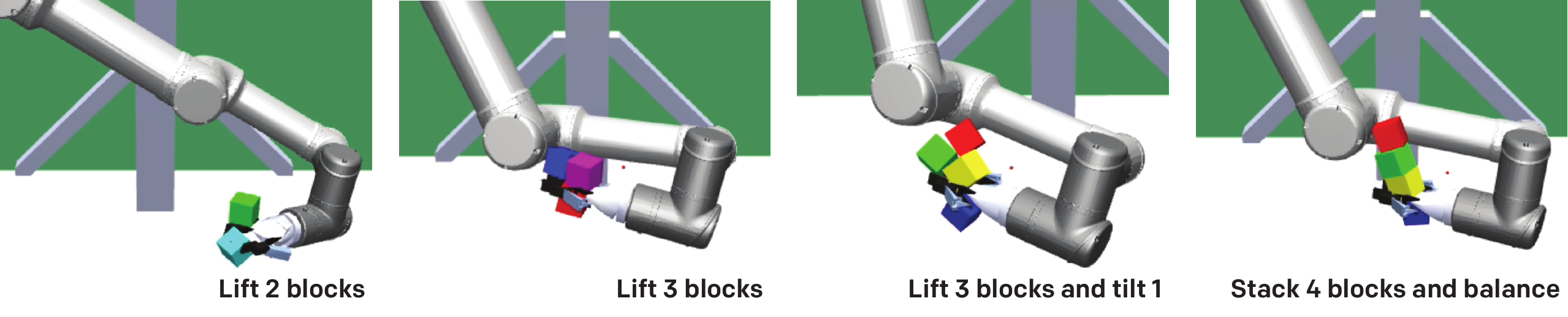}
\caption{\label{fig:exp-goal-example}Goals discovered by asymmetric self-play. Alice discovers many goals that are not covered by our manually designed holdout tasks on blocks.
\vspace{-0.35cm}
}
\end{figure}

\begin{SCfigure}[50][t!]
\includegraphics[width=7.0cm]{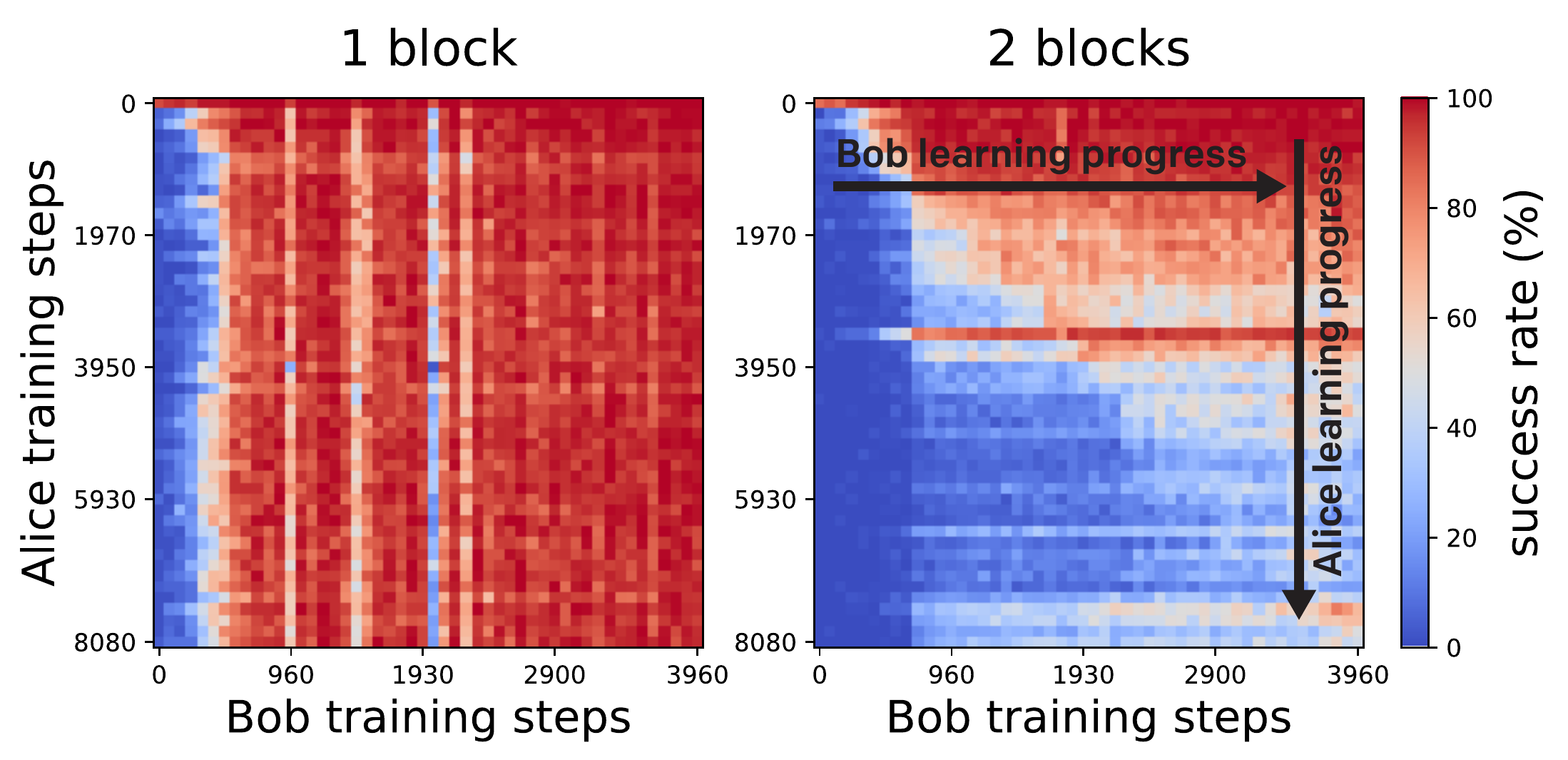}
\vspace{-0.5cm}
\caption{\label{fig:payoff}The empirical payoff matrix between Alice and Bob. Average success rate over multiple self-play episode is visualized. Alice with more training steps generates more challenging goals that Bob cannot solve yet. Bob with more training steps can achieve more goals against the same Alice.
}
\end{SCfigure}

Asymmetric self-play discovers novel goals and solutions that are not covered by our holdout tasks. As illustrated in \autoref{fig:exp-goal-example}, Alice can lift multiple blocks at the same time, build a tower and then keep it balanced using an arm joint. Although it is a tricky strategy for Bob to learn on its own, with ABC, Bob eventually acquires the skills for solving such complex tasks proposed by Alice. Videos are available at \url{https://robotics-self-play.github.io}.

\autoref{fig:payoff} summarizes Alice and Bob's learning progress against each other.
For every pair of Alice and Bob, we ran multiple self-play episodes and measured the success rate. 
We observe an interesting trend with 2 blocks.
As training proceeds, Alice tends to generate more challenging goals, where Bob shows lower success rate. With past sampling, Bob continues to make progress against versions of Alices from earlier optimization steps.
This visualization suggests a desired dynamic of asymmetric self-play that could potentially lead to unbounded complexity: Alice continuously generates goals to challenge Bob, and Bob keeps making progress on learning to solve new goals.

\subsection{Generalization to unseen objects and goals}
\label{sec:big-exp}

\begin{figure*}[h!]
    \centering
    \begin{subfigure}[b]{0.16\textwidth}
        \centering
        \includegraphics[width=1.0\textwidth]{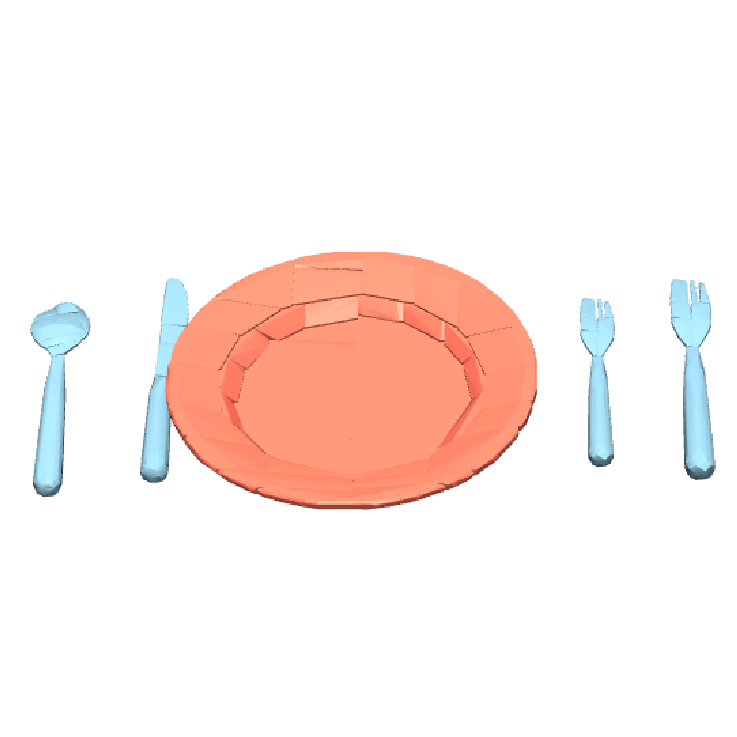}
        \caption{Table setting}
    \end{subfigure}
    \begin{subfigure}[b]{0.16\textwidth}
        \centering
        \includegraphics[width=1.0\textwidth]{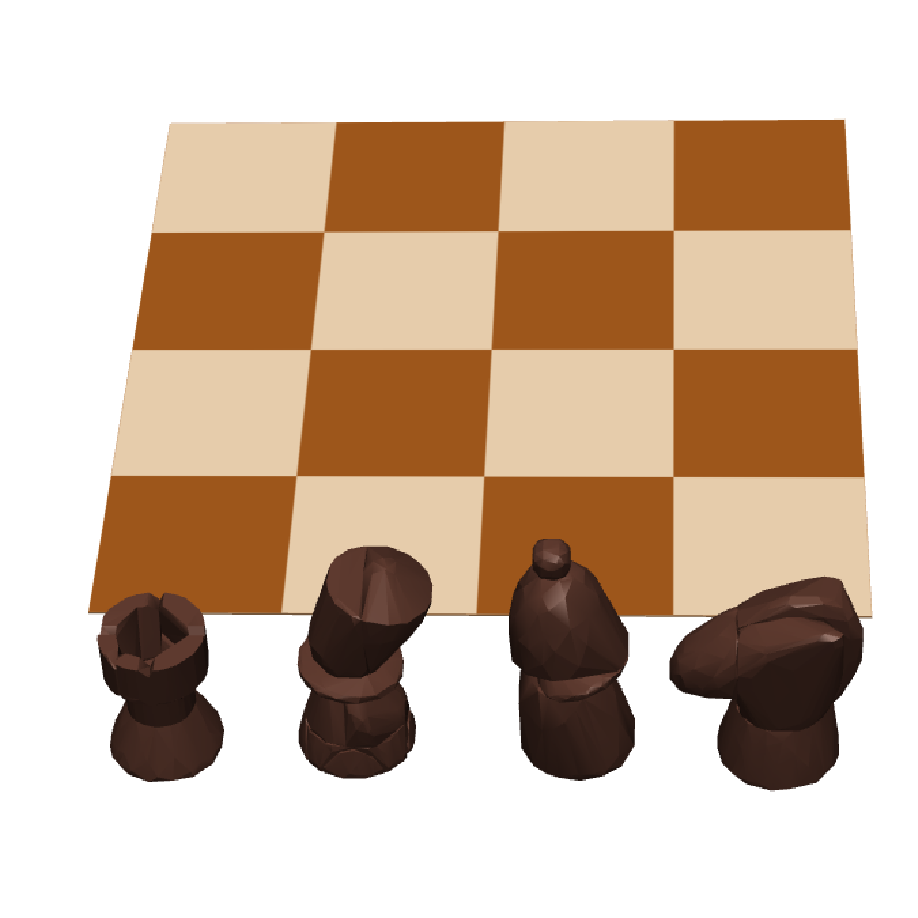}
        \caption{Mini chess}
    \end{subfigure}
    \begin{subfigure}[b]{0.16\textwidth}
        \centering
        \includegraphics[width=1.0\textwidth]{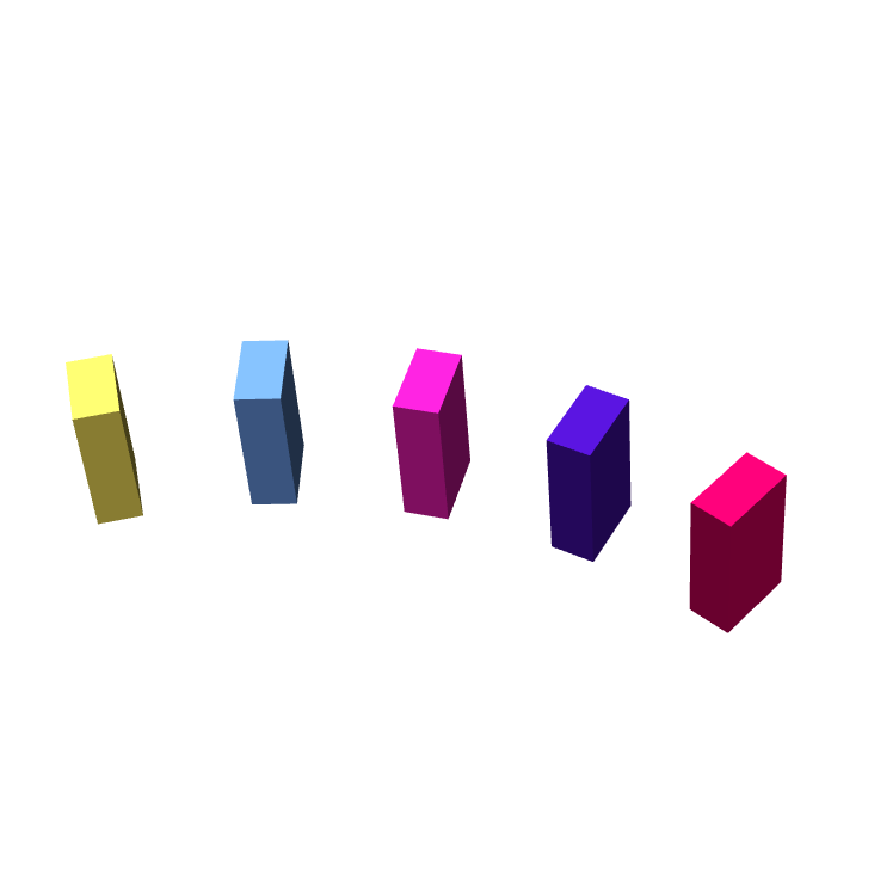}
        \caption{Domino}
    \end{subfigure}
    \begin{subfigure}[b]{0.16\textwidth}
        \centering
        \includegraphics[width=1.0\textwidth]{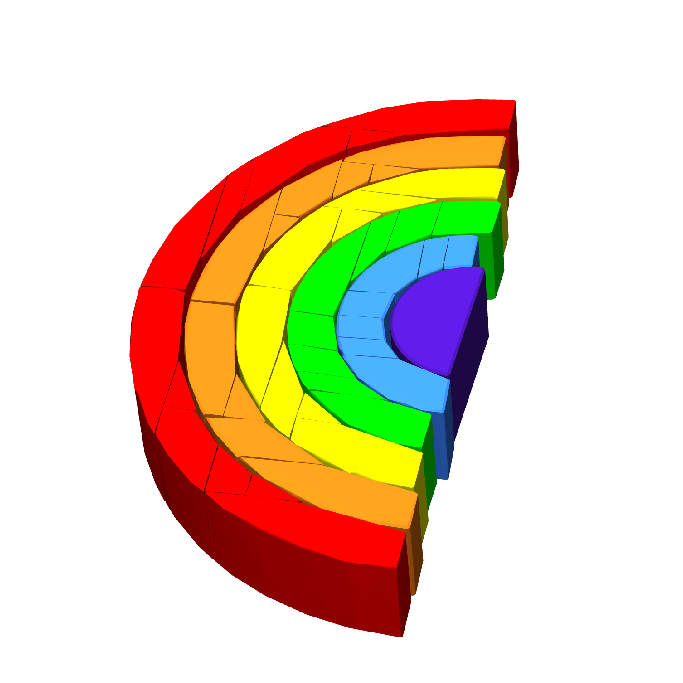}
        \caption{Rainbow}
    \end{subfigure}
    \begin{subfigure}[b]{0.16\textwidth}
        \centering
        \includegraphics[width=1.0\textwidth]{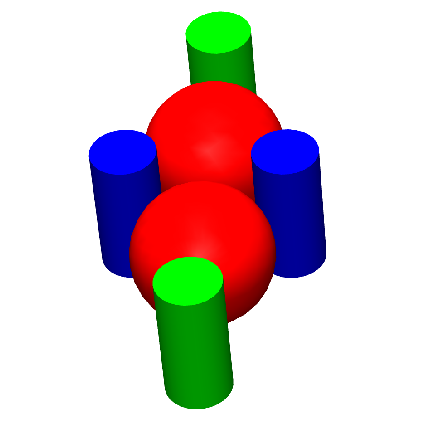}
        \caption{Ball-capture}
    \end{subfigure}
    \begin{subfigure}[b]{0.16\textwidth}
        \centering
        \includegraphics[width=1.0\textwidth]{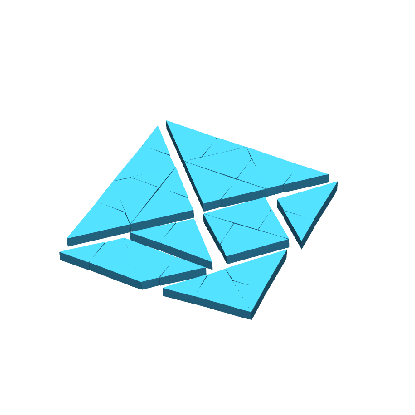}
        \caption{Tangram}
    \end{subfigure}
    \caption{\label{fig:complex-holdouts} Example holdout tasks involving unseen objects and complex goal states. The goal states are illustrated here, and the initial states have randomly placed objects.}
\end{figure*}

The experiments above show strong evidence that efficient curricula and novel goals can autonomously emerge in asymmetric self-play. To further challenge our approach, we scale it up to work with many more complex objects using more computational resources for training.
We train a hybrid policy in an environment containing up to 10 random ShapeNet~\citep{chang2015shapenet} objects. 
During training, we randomize the number of objects and the object sizes via Automatic Domain Randomization~\citep{openai2019solving}. 
The hybrid policy uses vision observations to extract information about object geometry and size.
We evaluate the Bob policy on a more diverse set of manipulation tasks, including semantically interesting ones. Many tasks contain unseen objects and complex goals, as illustrated in \autoref{fig:complex-holdouts}.

\begin{figure}[t!]
\centering
\includegraphics[width=1.0\linewidth]{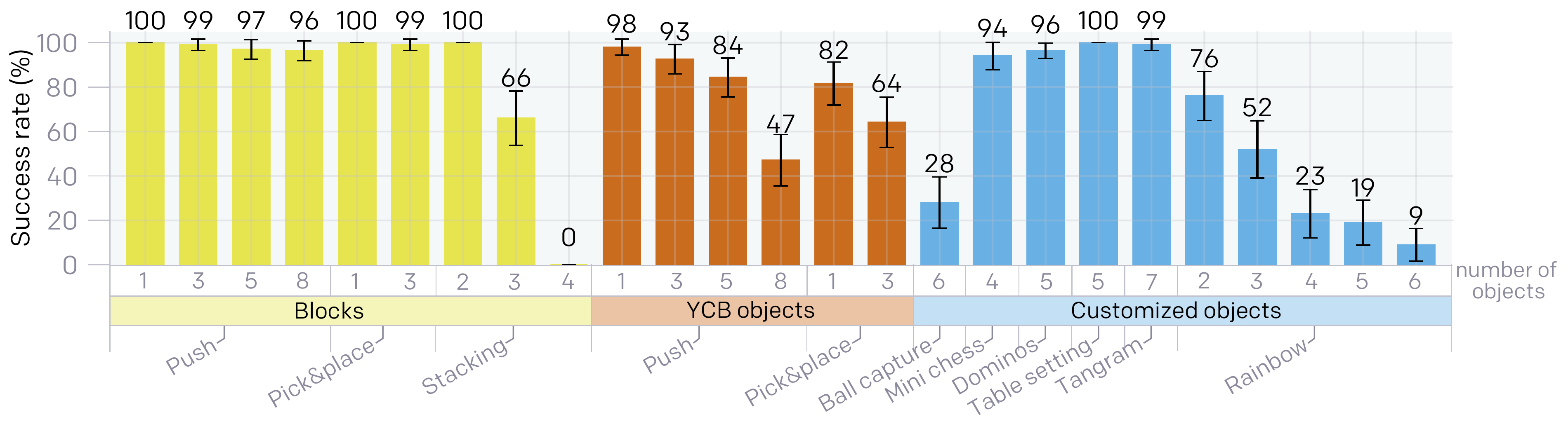}
\caption{\label{fig:big}Success rates of a single goal-conditioned policy solving a variety of holdout tasks, averaged over 100 trials. The error bars indicate the $99\%$ confidence intervals. Yellow, orange and blue bars correspond to success rates of manipulation tasks with blocks, YCB\protect\footnotemark objects and other uniquely built objects, respectively. Videos are available at \url{https://robotics-self-play.github.io}.
}
\end{figure}

The learned Bob policy achieves decent zero-shot generalization performance for many tasks. Success rates are reported in \autoref{fig:big}. 
Several tasks are still challenging. For example, \texttt{ball-capture} requires delicate handling of rolling objects and lifting skills. The \texttt{rainbow} tasks call for an understanding of concave shapes. Understanding the ordering of placement actions is crucial for stacking more than 3 blocks in the desired order.
The Bob policy learns such an ordering to some degree, but fails to fully generalize to an arbitrary number of stacked blocks.

\footnotetext{\url{https://www.ycbbenchmarks.com/object-models/}}

\subsection{Ablation studies}
\label{sec:exp-ablation}

\begin{figure}[h!]
\centering
\includegraphics[width=1.0\linewidth]{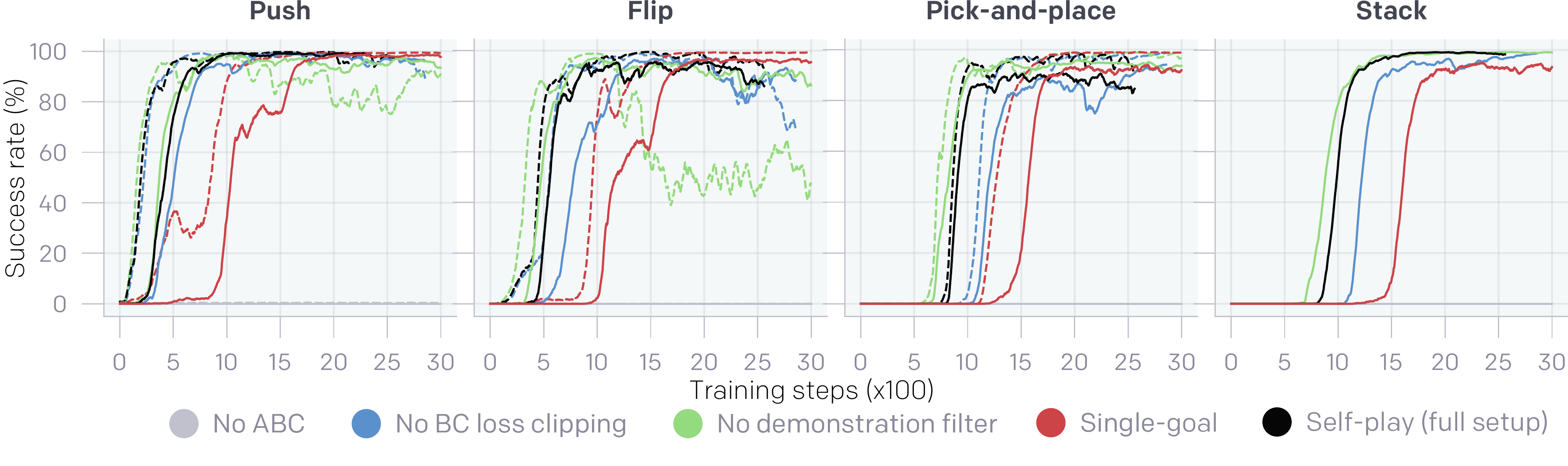}
\caption{\label{fig:abc_ablation} The ablation studies compare four ablation runs each with one component disabled with the full baseline. Solid lines are for 2-blocks, dashed lines are for 1-block.
The x-axis denotes the number of training steps via asymmetric self-play.
The y-axis is the zero-shot generalization performance of Bob policy at corresponding training steps.
}
\end{figure}

We present a series of ablation studies designed for measuring the importance of each component in our asymmetric self-play framework, including Alice behavioral cloning (ABC), BC loss clipping, demonstration filtering, and the multi-goal game setup. We disable a single ingredient in each ablation run and compare with the complete self-play baseline in \autoref{fig:abc_ablation}.

The \texttt{no ABC} baseline shows that Bob \textit{completely fails} to solve any holdout task without ABC, indicating that ABC is a critical mechanism in asymmetric self-play.
The \texttt{no BC loss clipping} baseline shows slightly slower learning on \texttt{pick-and-place} and \texttt{stack}, as well as some instabilities in the middle of training. Clipping in the BC loss is expected to help alleviate this instability by controlling the rate of policy change per optimizer iteration.
The \texttt{no demonstration filter} baseline shows noticeable instability on \texttt{flip}, suggesting the importance of excluding suboptimal demonstrations from behavioral cloning.
Finally, the \texttt{single-goal} baseline uses a single goal instead of 5 goals per episode during training. The evaluation tasks are also updated to require a single success per episode. Generalization of this baseline to holdout tasks turns out to be much slower and less stable. It signifies some advantages of using multiple goals per episode, perhaps due to the policy memory internalizing environmental information during multiple trials of goal solving.

The results of the ablation studies suggest that ABC with proper configuration and multi-goal game-play are critical components of asymmetric self-play, alleviating the importance of manual curricula and facilitating efficient learning.


\section{Conclusion}
\label{sec:conclusion}
One limitation of our asymmetric self-play approach is that it depends on a resettable simulation environment as Bob needs to start from the same initial state as Alice's. Therefore asymmetric self-play training has to happen in a simulator which can be easily updated to a desired state. 
In order to run the goal-solving policy on physical robots, we plan to adopt sim-to-real techniques in future work. Sim-to-real has been shown to achieve great performance on many robotic tasks in the real world~\citep{Sadeghi2017sim2real, tobin2017domain, james2019RCAN, openai2020block}. One potential approach is to pre-train two agents via asymmetric self-play in simulation, and then fine-tune the Bob policy with domain randomization or data collected on physical robots.

In conclusion, we studied asymmetric self-play as a framework for defining a single training distribution to learn many arbitrary object manipulation tasks.
Even without any prior knowledge about the target tasks, asymmetric self-play is able to train a strong goal-conditioned policy that can generalize to many unseen holdout tasks. 
We found that asymmetric self-play not only generates a wide range of interesting goals but also alleviates the necessity of designing manual curricula for learning such goals.
We provided evidence that using the goal setting trajectory as a demonstration for training a goal solving policy is essential to enable efficient learning. 
We further scaled up our approach to work with various complex objects using more computation, and achieved zero-shot generalization to a collection of challenging manipulation tasks involving unseen objects and unseen goals.

\clearpage

\clearpage
\subsubsection*{Author Contributions}

This manuscript is the result of the work of the entire OpenAI Robotics team. We list the contributions of every team member here grouped by topic and in alphabetical order.

\begin{itemize}

    \item Hyeonwoo Noh and Lilian Weng designed and implemented the asymmmetric self-play algorithm for robotics manipulation tasks.

    \item Vineet Kosaraju, Hyeonwoo Noh, Alex Paino, Matthias Plappert, Lilian Weng and Qiming Yuan developed the simulation environments for RL training.

    \item Casey Chu, Vineet Kosaraju, Alex Paino, Henrique Ponde de Oliveira Pinto, Qiming Yuan and Tao Xu worked on the vision stack configuration. 

    \item Ilge Akkaya, Ruben D'Sa, Arthur Petron and Raul Sampedro developed the robot controller and simulation.

    \item Qiming Yuan developed toolings for building holdout environments.

    \item Ilge Akkaya, Alex Paino, Arthur Petron, Henrique Ponde de Oliveira Pinto, Lilian Weng, Tao Xu and Qiming Yuan built a collection of the holdout environments.
    
    \item Henrique Ponde de Oliveira Pinto optimized the training infrastructure and developed monitoring metrics.

    \item Casey Chu, Hyeonwoo Noh and Lilian Weng drafted the manuscript.

    \item All authors refined and revised the manuscript.
    
    \item Ilge Akkaya, Alex Paino, Matthias Plappert, Peter Welinder, and Lilian Weng led aspects of this project and set research directions.
    
    \item Ilge Akkaya, Matthias Plappert and Wojciech Zaremba managed the team.

\end{itemize}

\subsubsection*{Acknowledgments}

We would like to thank Joel Lehman, Ingmar Kanitscheider and Harri Edwards for providing thoughtful feedback for the algorithm and earlier versions of the manuscript. We also would like to thank Bowen Baker and Lei Zhang for helpful discussion on the idea. Finally we are grateful for everyone at OpenAI, especially the Supercomputing team, for their help and support.

\clearpage

\bibliography{references}
\bibliographystyle{iclr2021_conference}

\clearpage
\appendix

\section{Asymmetric Self-play Game Setup}
\label{appendix:game}
\subsection{Goal validation}
\label{appendix:goal-validation}

Some of the goals set via asymmetric self-play may not be useful or interesting enough to be included in the training distribution.
For example, if Alice fails to touch any objects, Bob can declare a success without any action. A goal outside the table might be tricky to solve.
We label such goals as \textit{invalid} and make sure that Alice has generated a \textit{valid} goal before starting Bob's turn.

Even when Alice generates a valid goal, we can still penalize certain undesired goals. 
Specifically, if the visual perception is limited to a restricted area on the table due to the camera setting, we can penalize goals containing objects outside that range of view.

For goal validation, we check in the following order:
\begin{enumerate}
    \item We check whether any object has moved. If not, the goal is considered \textit{invalid} and the episode resets.
    \item We check whether all the objects are on the table. If not, the goal is considered \textit{invalid} and the episode resets.
    \item We check whether a valid goal has objects outside the placement area, defined to be a 3D space that the robot end effector can reach and the robot camera can see. If any object is outside the area, the goal is deemed valid but obtains a \emph{out-of-zone} penalty reward and the episode continues to switch to Bob's turn.
\end{enumerate}

\subsection{Reward structure}
\label{appendix:reward}

\autoref{table:reward} shows the reward structure for a single turn for goal setting and solving. The reward for Alice is based on whether it successfully generates a valid goal, and whether the generated goal is solved by Bob.
Alice obtains 1 point for a valid goal, and obtains an additional 5 point game reward if Bob fails to achieve it. Additionally, a goal out of placement area triggers a $-3$ penalty.
Rewarding Alice based only on Bob's success or failure is simpler than the original reward from \cite{sukhbaatar2017intrinsic}, but we didn't notice any degradation from this simplification (Appendix~\ref{appendix:baselines-goal}).

Since Bob is a goal-conditioned policy, we provide \emph{sparse} goal-conditioned rewards. Whenever one object is placed at its desired position and orientation, Bob obtains 1 point per-object reward. Bob obtains -1 reward such that the sum of per-object reward is at most 1 during a given turn.  When all the objects are in the goal state, Bob obtains a 5 point success reward and its turn terminates. If Bob reaches a maximum number of allowed steps before achieving the goal, it is deemed a failure with 0 point reward.

When checking whether a goal has been achieved, we compare the position and orientation of each object with its goal position and orientation. For position, we compute the Euclidean distance between object centers. For rotation, we represent the orientation of an object by three Euler angles on three dimensions, roll, pitch, and yaw, respectively, and we compute the minimum angle needed to rotate the object into the goal orientation. If the distance and angle for all objects are less than a small error (0.04 meters and 0.2 radians respectively), we consider the goal achieved.

\begin{table}[h!]
\centering
\caption{\label{table:reward}Reward structure for a single goal.}
\begin{tabular}{ c c c c}
\hline
Alice & Bob & Alice reward & Bob reward \\ 
\hline
Invalid goal & - & 0 & - \\ 
Out-of-zone goal & Failure & 1 - 3 + 5 & 0 + per-object reward \\ 
Out-of-zone goal & Success & 1 - 3 + 0 & 5 + per-object reward \\ 
Valid goal & Failure & 1 + 5 & 0 + per-object reward \\ 
Valid goal & Success & 1 + 0 & 5 + per-object reward \\
\hline
\end{tabular}
\end{table}

\subsection{Multi-goal game structure}
\label{appendix:multi-goal-game}

\begin{figure}[h!]
\centering
\includegraphics[width=0.4\linewidth]{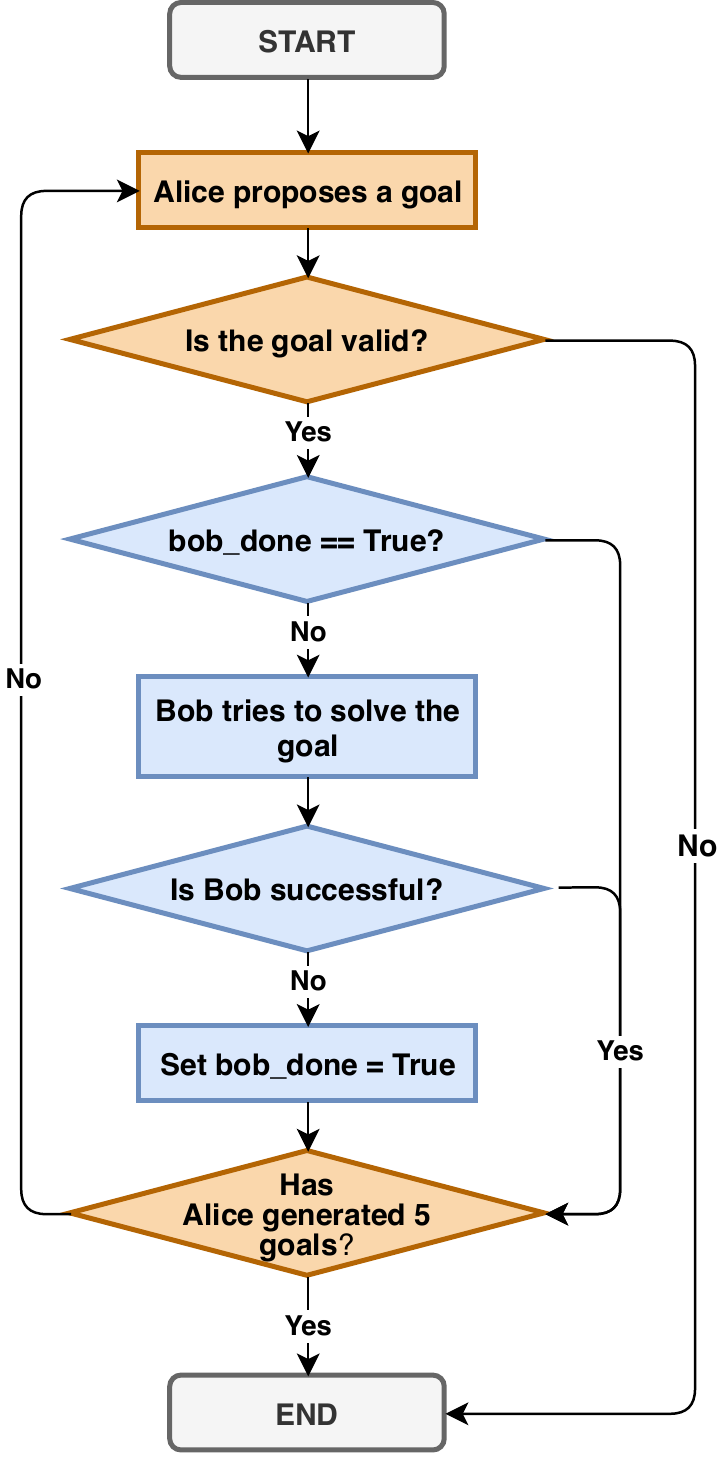}
\caption{\label{fig:flowchart}The flow chart of asymmetric self-play with a multi-goal game structure. The steps in orange belong to Alice while the blue ones belong to Bob.}
\end{figure}

The overall flowchart of asymmetric self-play with a multi-goal structure is illustrated in \autoref{fig:flowchart}.

We expect a multi-goal game structure to induce a more complicated goal distribution, as goals can be built on top of each other. For example, in order to stack 3 blocks, you might need to stack 2 blocks first as a subgoal at the first step. 
A multi-goal structure also encourages Bob to internalize environmental information during multiple trials of goal solving. Many aspects of the environment, such as the simulator's physical dynamics and properties of objects, stay constant within one episode, so Bob can systematically investigate these constant properties and exploit them to adapt its goal solving strategy accordingly. Similar behavior with multi-goal setting was observed by~\citet{openai2019solving}.

In our experiments, when we report the success rate from multi-goal episodes, we run many episodes with a maximum of 5 goals each and compute the success rate as 

\begin{equation*}
    \texttt{success\_rate} = \frac{\texttt{total\_successes}}{\texttt{total\_goals}}.
\end{equation*}

Note that because each episode terminates after one failure, at the end of one episode, we would have either $\texttt{total\_goals} = \texttt{total\_successes}$ if Bob succeeded at every goal, or $\texttt{total\_goals} = \texttt{total\_successes} + 1$ if Bob fails in the middle.

\subsection{Training algorithm}
\label{appendix:train-algorithm}

Algorithm~\ref{algo:train-selfplay} and \ref{algo:update-buffer} describe pseudocode for the training algorithm using asymmetric self-play. 
Both policies are optimized via Proximal Policy Optimization (PPO)~\citep{schulman2017ppo}.
Additionally, Bob optimizes the Alice behavioral cloning (ABC) loss using Alice's demonstrations collected during the interplay between two agents. 
In the algorithm, $\mathcal{L}_\text{RL}$ denotes a loss function for PPO and $\mathcal{L}_\text{ABC}$ denotes a loss function for ABC.
A trajectory $\tau$ contains a list of (state, action, reward) tuples, $\tau=\{(s_0, a_0, r_0), (s_1, a_1, r_1), \dots \}$.
A goal-augmented trajectory $\tau_\text{BC}$ contains a list of (state, goal, action, reward) tuples, $\tau_\text{BC}=\{(s_0, g, a_0, r_0), (s_1, g, a_1, r_1), \dots \}$.


\begin{algorithm}[t!]
\caption{\textbf{Asymmetric self-play}}
\label{algo:train-selfplay}
\begin{algorithmic}
\Require $\theta_\text{A}, \theta_\text{B}$ \Comment{Initial parameters for Alice and Bob}
\Require $\eta$ \Comment{RL learning rate}
\Require $\beta$ \Comment{weight of BC loss}
\For{training steps $= 1, 2, ...$}
    \State $\theta_\text{A}^\text{old} \leftarrow \theta_\text{A}, \theta_\text{B}^\text{old} \leftarrow \theta_\text{B}$ \Comment{initialize behavior policy parameters}
    \For{each rollout worker} \Comment{parallel data collection}
        \State $\mathcal{D}_\text{A}, \mathcal{D}_\text{B}, \mathcal{D}_\text{BC}$ $\leftarrow$ $\textbf{CollectRolloutData}(\theta_\text{A}^\text{old}, \theta_\text{B}^\text{old})$ \Comment{replay buffers for Alice, Bob and ABC}
    \EndFor
    \State $\theta_\text{A} \leftarrow \theta_\text{A} - \eta \nabla_{\theta_\text{A}}\mathcal{L}_\text{RL}$ \Comment{optimize PPO loss with data popped from $\mathcal{D}_\text{A}$}
    \State $\theta_\text{B} \leftarrow \theta_\text{B} - \eta \nabla_{\theta_\text{B}}\big[\mathcal{L}_\text{RL} + \beta \mathcal{L}_\text{ABC}\big]$ \Comment{optimize RL loss with $\mathcal{D}_\text{B}$ and ABC loss with $\mathcal{D}_\text{BC}$}
\EndFor
\end{algorithmic}
\end{algorithm}

\begin{algorithm}[t!]
\caption{\textbf{CollectRolloutData}}
\label{algo:update-buffer}
\begin{algorithmic}
\Require $\theta^\text{old}_\text{A}, \theta^\text{old}_\text{B}$ \Comment{behavior policy parameters for Alice and Bob}
\Require $\pi_\text{A}(a|s; \theta^\text{old}_\text{A}), \pi_\text{B}(a|s, g; \theta^\text{old}_\text{B})$ \Comment{policies for Alice and Bob}
\Require $\xi$ \Comment{whether Bob succeeded to achieve a goal}
\State $\mathcal{D}_\text{A} \gets \varnothing, \mathcal{D}_\text{B} \gets \varnothing, \mathcal{D}_\text{BC} \gets \varnothing$ \Comment{Initialize empty replay buffers.}
\State $\xi \leftarrow$True  \Comment{initialize to True (success)}
\For{number of goals = 1, ..., 5}
    \State $\tau_\text{A}, g \leftarrow \textbf{GenerateAliceTrajectory}(\pi_A, \theta^\text{old}_\text{A})$ \Comment{generates a trajectory $\tau_\text{A}$ and a goal $g$}
    \If{goal $g$ is invalid}
        \State break
    \EndIf
    \If{$\xi$ is True}
        \State $\tau_\text{B}, \xi \leftarrow \textbf{GenerateBobTrajectory}(\pi_\text{B}, \theta^\text{old}_\text{B}, g)$ \Comment{generate a trajectory $\tau_\text{B}$ and update $\xi$}
        \State $\mathcal{D}_\text{B} \leftarrow \mathcal{D}_\text{B} \cup \{\tau_\text{B}\}$ \Comment{update replay buffer for Bob}
    \EndIf
    \State $r_\text{A} \leftarrow \textbf{ComputeAliceReward}(\xi, g)$
    \State $\tau_\text{A}[-1][2] \leftarrow r_\text{A}$  \Comment{overwrite the last reward in trajectory $\tau_\text{A}$ with $r_\text{A}$}
    \State $\mathcal{D}_\text{A} \leftarrow \mathcal{D}_\text{A} \cup \{\tau_\text{A}\}$ \Comment{update replay buffer for Alice}
    \If{$\xi$ is False}
        \State $\tau_\text{BC} \leftarrow \textbf{RelabelDemonstration}(\tau_\text{A}, g, \pi_\text{B}, \theta_\text{B}^\text{old})$  \Comment{relabeled to be goal-augmented}
        \State $\mathcal{D}_\text{BC} \leftarrow \mathcal{D}_\text{BC} \cup \{\tau_\text{BC}\}$ \Comment{update replay buffer for ABC}
    \EndIf
\EndFor
\State \Return $\mathcal{D}_\text{A}, \mathcal{D}_\text{B}, \mathcal{D}_\text{BC}$
\end{algorithmic}
\end{algorithm}

\section{Training Setup}
\label{appendix:setup}
\subsection{Simulation Setup}
\label{appendix:sim}

We utilize the MuJoCo physics engine~\citep{mujoco} to simulate our robot environment and render vision observations and goals. We model a UR16e robotic arm equipped with a RobotIQ 2F-85 parallel gripper end effector. The robot arm is controlled via its tool center point (TCP) pose that is actuated via MuJoCo constraints. Additionally, we use a PID controller to actuate the parallel gripper using position control.

\subsection{Action Space}
\label{appendix:action_space}
We define a 6-dimensional action space consisting of 3D relative gripper position, 2D relative gripper rotation, and a 1D desired relative gripper finger position output that is applied symmetrically to the two gripper pinchers. The two rotational degrees of freedom correspond to yaw and pitch axes (wrist rotation and wrist tilt) respectively, with respect to the gripper base. We use a discretized action space with 11 bins per dimension and learn a multi-categorical distribution.

\subsection{Observation Space}
\label{appendix:obs_space}

We feed observations of robot arm position, gripper position, object state, and goal state into the policy. 
The object state observation contains each object's position, rotation, velocity, rotational velocity, the distance between the object and the gripper, as well as whether this object has contacts with the gripper.
The goal state observation includes each object's desired position and rotation, as well as the relative distance between the current object state and the desired state.

\begin{figure*}[t!]
    \centering
    \begin{subfigure}[b]{0.32\textwidth}
        \centering
        \includegraphics[width=1.0\textwidth]{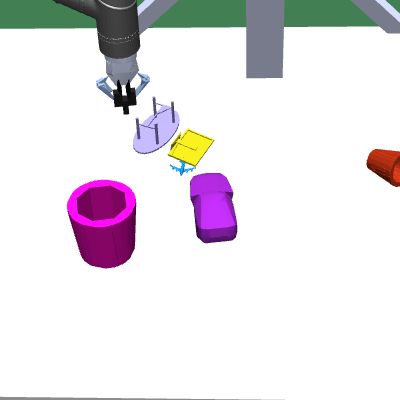}
        \caption{Front camera}
    \end{subfigure}
    \begin{subfigure}[b]{0.32\textwidth}
        \centering
        \includegraphics[width=1.0\textwidth]{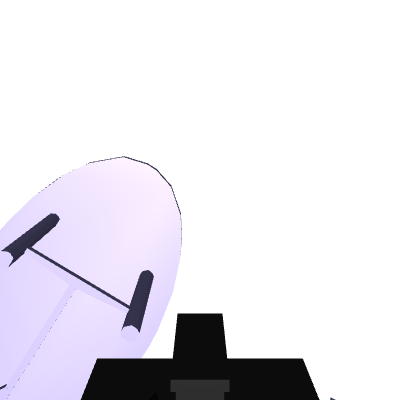}
        \caption{Wrist camera}
    \end{subfigure}
    \begin{subfigure}[b]{0.32\textwidth}
        \centering
        \includegraphics[width=1.0\textwidth]{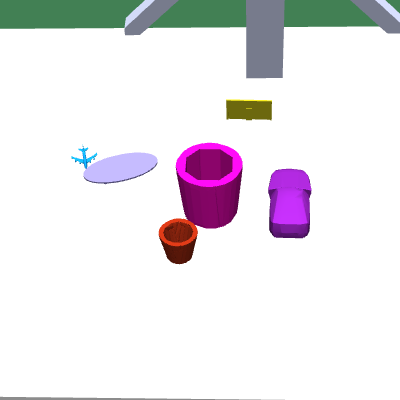}
        \caption{Front camera (goal)}
    \end{subfigure}
    \caption{\label{fig:vision-obs} Example vision observations from our camera setup. (a) observation from a camera mounted in front of the table (the front camera). (b) observation from the mobile camera mounted on the gripper wrist. (c) goal observation from the front camera.}
\end{figure*}

In the hybrid policy for the ShapeNet training environment, we additionally feed three camera images into the policy: an image of the current state captured by a fixed camera in front of the table, an image of the current state from a camera mounted on the gripper wrist, and an image of the goal state from the fixed camera. Figure~\ref{fig:vision-obs} illustrates the example observations from our camera setup. Both Alice and Bob take robot and object state observations as inputs, but Alice does not take goal state inputs since it is not goal-conditioned.

\subsection{Model Architecture}
\label{appendix:model_arch}

We use independent policy and value networks in the PPO policy. Both have the same observation inputs and network architecture, as illustrated in \autoref{fig:model_arch}. The permutation invariant embedding module concatenates all the observations per object, learns an embedding vector per object and then does max pooling in the object dimension.
The vision module uses the same model architecture as in IMPALA~\citep{espeholt2018impala}. For all experiments, we use completely separate parameters for the policy and the value network except the vision module, which is shared between them.

\begin{figure}[h!]
\centering
\includegraphics[width=\linewidth]{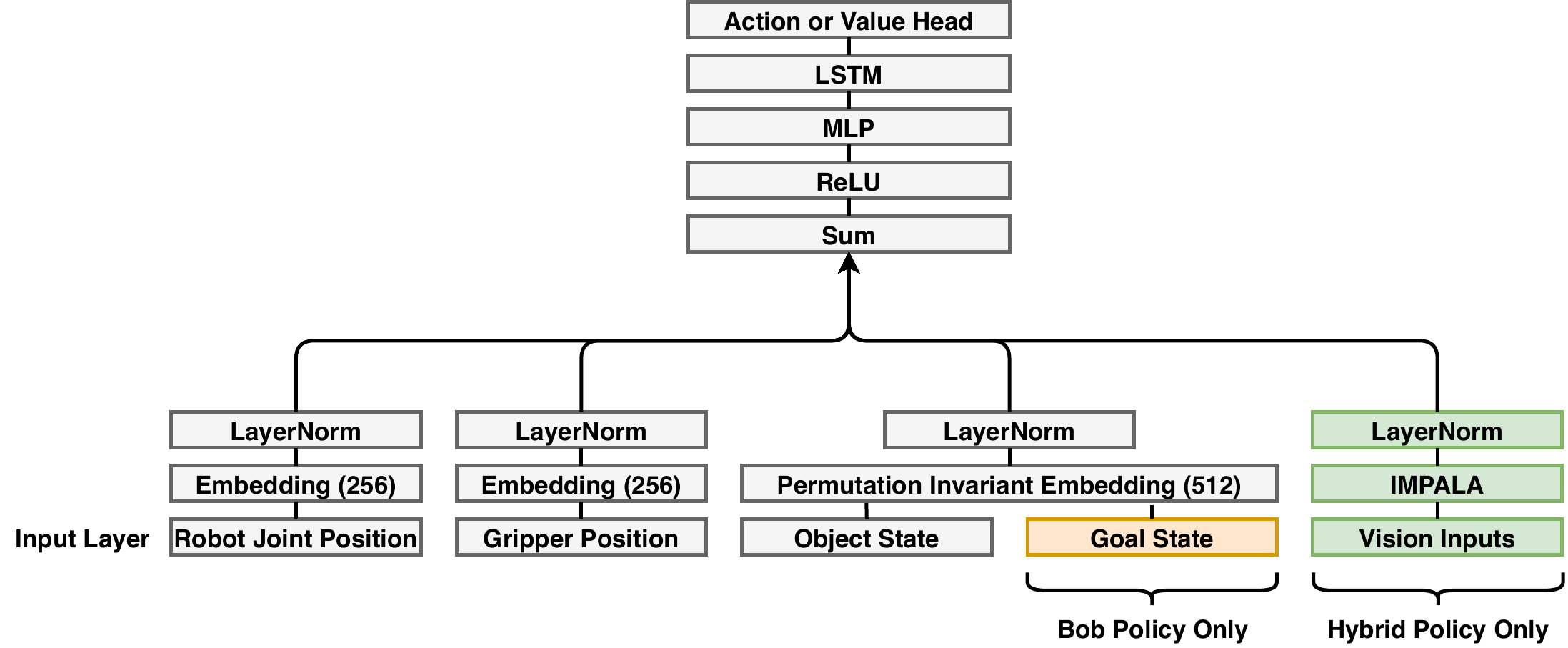}
\caption{\label{fig:model_arch}Network architecture of value/policy network.}
\end{figure}

\subsection{Hyperparameters}

Hyperparameters used in our PPO policy and asymmetric self-play setup are listed in \autoref{table:ppo_hparams} and \autoref{table:hardware}. 

\begin{table}[h!]
    \footnotesize
    \centering
    \caption{Hyperparameters used for PPO.}
    \renewcommand{\arraystretch}{1.3}
    \begin{tabular}{@{}ll@{}}
        \toprule
        \textbf{Hyperparameter} & \textbf{Value} \\ 
        \midrule
        discount factor $\gamma$ & $0.998$ \\
        Generalized Advantage Estimation (GAE) $\lambda$ & $0.95$ \\
        entropy regularization coefficient & $0.01$ \\
        PPO clipping parameter $\epsilon_\text{ppo}$ & $0.2$ \\
        ABC clipping parameter $\epsilon$ & $0.2$ \\
        optimizer & Adam~\citep{Kingma2014AdamAM} \\
        learning rate $\eta$ & $3 \times 10^{-4}$  \\
        sample reuse (experience replay) & 3 \\
        value loss weight & $1.0$ \\
        ABC loss weight & $0.5$ \\
    \bottomrule
    \end{tabular}
    \label{table:ppo_hparams}
\end{table}

The maximum goal solving steps for Bob reported in Table~\ref{table:hardware} is the number of steps allowed per object within one episode. If Bob has spent all these time steps but still cannot solve the goal, it deems a failure and the episode terminates for Bob.

\begin{table}[h!]
    \footnotesize
    \centering
    \caption{Hyperparameters used for hardware configuration, batch size and self-play episode length.}
    \renewcommand{\arraystretch}{1.3}
    \begin{tabular}{@{}lll@{}}
        \toprule
        \textbf{Hyperparameter} & \textbf{1-2 Block manipulation} (state) & \textbf{ShapeNet object rearrangement} (hybrid) \\ 
        \midrule
        GPUs per policy & $1$ & $32 \times 8$ \\
        rollout worker CPUs & $64 \times 29$ & $576 \times 29$ \\
        batch size& $4096$ & $55 \times 32 \times 8$ \\
        Alice's goal setting steps $T$ & $100$ & $250$ \\
        Bob's maximum goal solving steps & $200$ & $600$ \\
    \bottomrule\end{tabular}
    \label{table:hardware}
\end{table}

\FloatBarrier

\subsection{Holdout Tasks}
\label{appendix:holdouts}

Here are a list of tasks for evaluating zero-shot generalization capability of the hybrid policy. Some of them are visualized in \autoref{fig:big}. Note that none of the objects here appear in the training data.

\begin{itemize}
    \item Table setting: arrange a table setting consisting of a plate, spoon, knife, salad fork, and main course fork.
    \item Mini chess: place four chess pieces next to a chess board.
    \item Rainbow (2-6 pieces): build a rainbow out of colored, wooden pieces by placing the half-circle shapes together so they resemble a rainbow. We have 5 variations of the rainbow tasks by taking different numbers of pieces which are indexed from the outer circle to inner one.
    \item Ball-capture: capture two, red field-hockey balls by placing four (two blue and two green) cylinders at so their rotational axes intersect rays from the spheres at roughly 90, 240, and 300 degree angles about the same axial (Z) direction.
    \item Tangram Puzzle: move blue pieces to form the standard, seven piece tangram square solution.
    \item Domino: stand up 5 wooden domino pieces in a curved layout.
    \item Block push (1--8 objects): push blocks to match position and orientation of goal configurations. All goal objects are on the surface of the table.
    \item Block pick-and-place (1--3 objects): push blocks, and lift up one block in the air. One goal object is in the air and all other goal objects are on the table surface.
    \item Block stacking (2--4 objects): stack blocks to form a tower in a specific location and specific rotation.
    \item YCB object push (1--8 objects): push YCB objects\footnote{\url{https://www.ycbbenchmarks.com/object-models/}} to match position and orientation of goal configurations. All goal objects are on the surface of the table.
    \item YCB pick-and-place (1--3 objects): push YCB objects and lift up one YCB object in the air. One goal object is in the air and all other goal objects are on the table surface.
\end{itemize}

By default, each holdout task presents 5 goals per episode and terminates episodes upon a failure. 
Exceptions are Table setting, Mini chess, Rainbow (2-6 pieces), Ball-capture, and Tangram, which only present a single goal per episode because only a single fixed goal configuration is available for each holdout task.

\section{Non Self-play Baselines}
\label{appendix:baselines}
\subsection{Baselines for curriculum}
\label{appendix:baselines-curriculum}

We compared asymmetric self-play with several baselines incorporating hand-designed curricula in Sec.~\ref{sec:exp-curriculum}.

All the baselines are trained on a mixture of push, flip, pick-and-place, and stacking tasks as the goal distribution. The initial state of objects is generated by randomly placing objects within the placement area of the table without overlaps. The number of objects is sampled from $\{1, 2\}$ with equal probability.

Factorized Automatic Domain Randomization (FADR)~\citep{openai2019solving} is applied to grow curriculum parameters described below. Precisely, for each parameter we track a list of performance scores when the parameter is configured at current maximum and other parameters are randomly sampled. The value of this parameter will be increased if the tracked score rises above a threshold.

\begin{enumerate}
    \item The \texttt{no curriculum} baseline trains the goal-conditioned policy directly on a fixed goal distribution and environment parameters. Precisely there are $50\%$ goals for push and flip, $35\%$ for pick-and-place, and $15\%$ for stacking.

    \item The \texttt{curriculum:distance} baseline uses a hand-designed curriculum over (1) \texttt{goal\_distance\_ratio}: the Euclidean distance between the initial position of the objects and their goal positions, and (2) \texttt{goal\_rotation\_weight}: the weight applied on the object rotation distance for goal state matching. 
    At the beginning of the episode, given an initial position $x_0$ and a goal position $x_g$, we artificially make the goal easier according to \texttt{goal\_distance\_ratio} by resetting the goal position to $x_g' = x_0 + (x_g - x_0) \times \mathtt{goal\_distance\_ratio}$. A small ratio reduces the distance and thus makes the task less difficult.
    The parameter \texttt{goal\_rotation\_weight} controls how much we care about a good match between object rotation. Given the current object rotation $r_t$ and a desired rotation $r_g$, we consider them as a valid match if $\vert r_t - r_0 \vert \times \mathtt{goal\_rotation\_weight} < r_\text{threshold}$, where $r_\text{threshold} = 0.2$ (radians) is the success threshold for rotational matching. In other words, a small goal rotation weight creates a less strict success threshold.
    Both parameters range between $[0, 1]$ and gradually increase from $0$ to $1$ as training progresses.
    
    \item The \texttt{curriculum:distribution} controls the proportion of pick-and-place and stacking goals via two ADR parameters, \texttt{pickup\_proba} and \texttt{stack\_proba}. When sampling new goals, with probability \texttt{pickup\_proba}, a random object is moved up to the air and with probability \texttt{stack\_proba}, we consider a small 2-block tower as the goal.
    Both parameters range between $[0, 0.5]$ and gradually increase from $0$ to $0.5$ as training progresses.
    
    \item The \texttt{curriculum:full} baseline adopts all the ADR parameters described so far, \texttt{goal\_distance\_ratio}, \texttt{goal\_rotation\_weight}, \texttt{pickup\_proba} and \texttt{stack\_proba}. When setting up a pick-and-place goal, the height above the table surface is also interpolated according to \texttt{goal\_distance\_ratio}.

\end{enumerate}

\subsection{Comparison with timestep-based reward}
\label{appendix:baselines-goal}

Contrary to timestep-based reward originally proposed by \cite{sukhbaatar2017intrinsic}, we reward Alice simply based on the success or failure of Bob's goal solving attempt.
We compare our simplified reward structure with timestep-based reward for Alice as described in~\citet{sukhbaatar2017intrinsic} with time reward scale factor $0.01$ ($\gamma=0.01$ based on the notation from \cite{sukhbaatar2017intrinsic}). The performance is very similar.
Note that this result is not a direct comparison to~\citet{sukhbaatar2017intrinsic}, but an ablation study of two different reward functions based on our best asymmetric self-play configuration.

\begin{figure}[h!]
\centering
\includegraphics[width=\linewidth]{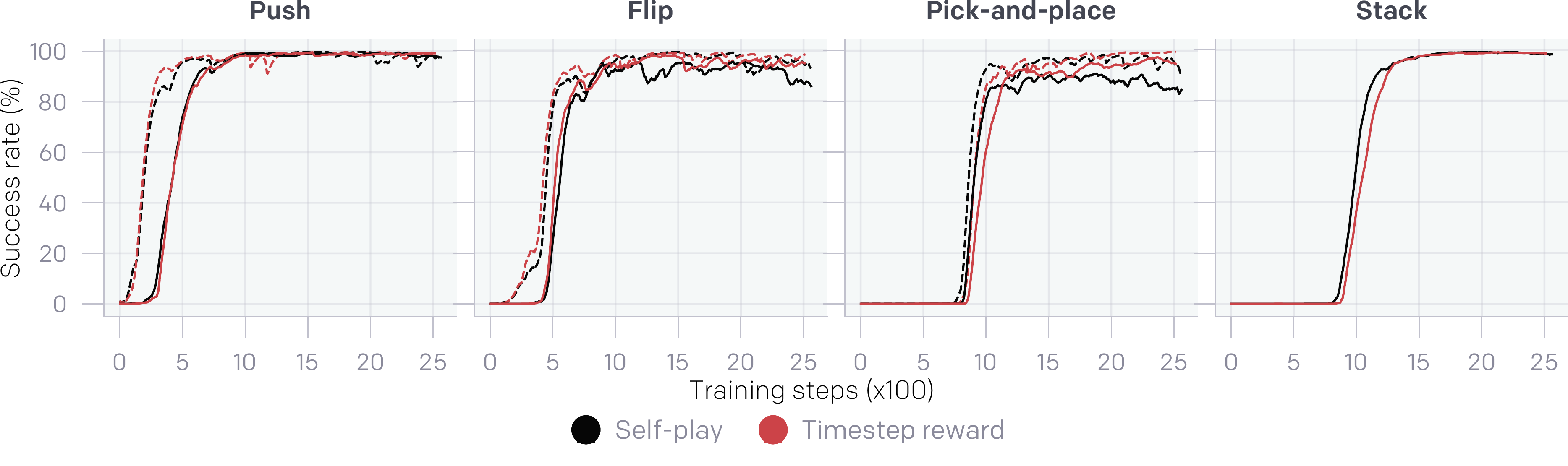}
\caption{\label{fig:model_arch}The comparison of our asymmetric self-play reward with the timestep-based reward.}
\end{figure}

\end{document}